\documentclass[conference]{acmsiggraph}

\TOGonlineid{45678}
\TOGvolume{0}
\TOGnumber{0}
\TOGarticleDOI{1111111.2222222}
\TOGprojectURL{}
\TOGvideoURL{}
\TOGdataURL{}
\TOGcodeURL{}

\usepackage[bookmarks=false]{hyperref}
\usepackage{color}
\usepackage{graphicx}
\usepackage{subfigure}
\renewcommand{\vec}[1]{\mathbf{#1}}
\usepackage{algorithm}
\usepackage{algpseudocode}
\makeatletter

\newcommand{\Rmnum}[1]{\expandafter\@slowromancap\romannumeral #1@}
\makeatother

\title{SeqFace: Make full use of sequence information for face recognition}

\author{
\vspace{1mm} Wei Hu$^1$            \thanks{e-mail: huwei@mail.buct.edu.cn}
\hspace{8mm} Yangyu Huang$^2$
\hspace{8mm} Fan Zhang$^1$
\hspace{8mm} Ruirui Li$^1$
\hspace{8mm} Wei Li$^1$
\hspace{8mm} Guodong Yuan$^2$
\\
\begin{minipage}{8.5cm} \center College of Information Science and Technology, \\ Beijing University of Chemical Technology, China \end{minipage}\hspace{0.2cm}
\begin{minipage}{8.5cm} \center YUNSHITU Corp., China \end{minipage}\hspace{0.2cm}
\vspace{-3mm} }
\pdfauthor{Wei Hu}


\begin{document}

%% \teaser{
%%   \includegraphics[height=1.5in]{images/sampleteaser}
%%   \caption{Spring Training 2009, Peoria, AZ.}
%% }

\maketitle

\begin{abstract}
Deep convolutional neural networks (CNNs) have greatly improved the Face Recognition (FR) performance in recent years. Almost all CNNs in FR are trained on the carefully labeled datasets containing plenty of identities. However, such high-quality datasets are very expensive to collect, which restricts many researchers to achieve state-of-the-art performance. In this paper, we propose a framework, called SeqFace, for learning discriminative face features. Besides a traditional identity training dataset, the designed SeqFace can train CNNs by using an additional dataset which includes a large number of face sequences collected from videos. Moreover, the label smoothing regularization (LSR) and a new proposed discriminative sequence agent (DSA) loss are employed to enhance discrimination power of deep face features via making full use of the sequence data. Our method achieves excellent performance on Labeled Faces in the Wild (LFW), YouTube Faces (YTF), only with a single ResNet. The code and models are publicly available onlien\footnote{https://github.com/huangyangyu/SeqFace}.
\end{abstract}

%\begin{CRcatlist}
%  \CRcat{I.3.3}{Computer Graphics}{Three-Dimensional Graphics and Realism}{Display Algorithms}
%  \CRcat{I.3.7}{Computer Graphics}{Three-Dimensional Graphics and Realism}{Radiosity};
%\end{CRcatlist}

\keywordlist

%% Use this only if you're preparing a technical paper to be published in the
%% ACM 'Transactions on Graphics' journal.

\TOGlinkslist

%% Required for all content.

\copyrightspace

\section{Introduction}
\label{sec:Introduction}

In most scenario, FR is a metric learning problem since it is impossible to classify faces to known identities in the training set. Recently, deep CNNs are widely used in FR, due to their great discriminative feature learning capability. The face feature is mainly trained via two types of methods according to their loss functions in CNN models. One method uses classification loss functions, such as softmax loss~\cite{deepface,frbyclass1,frbyclass2}. The other type uses metric learning loss functions, such as contrastive loss and triplet loss~\cite{tripletloss,contrastiveloss}. In many recent CNNs for FR, two types of loss functions are usually combined together for learning face features~\cite{centerloss}. All these loss functions aim to maximize the inter-identity variations and minimize the intra-identity variations under a certain metric space. No matter which loss functions are applied, we find that the training data share the same type, called \emph{identity data} in the paper.

An identity dataset includes $M$ faces of $N$ identities, and each face in the dataset is clearly labeled as the image of the $i$-th ($0 \leq i < N$) identity. Currently all public or private datasets for training deep face features, such as CASIA~\cite{casia}, MS-Celeb-1M~\cite{msceleb} and CelebFaces~\cite{deepid2}, belong to identity datasets. However, a large-scale high-quality identity dataset is very expensive to construct, since it could cost lots of effort and money. \emph{Identity data} need two kinds of information: face image and identity annotation. Identities in most public and private datasets are celebrities, because celebrity photos are rather easily crawled and annotated from the Internet. However, a celebrity dataset might be not a satisfied training dataset, if there are obvious differences between the evaluated faces and the celebrity faces in age, race, pose, and so on.

Beside photos from the Internet, videos (movies, TVs, surveillance videos, etc.) can also provide large quantities of face images, but few works utilize these face images because labelling process of identities is relatively difficult. However, it might be necessary to collect such faces as training data in some circumstances, such as surveillance. Face detection and tracking on videos can automatically generate data with lots of face sequences, and each sequence contains several faces of one identity. We call this type of data \emph{sequence data}.
\begin{figure}[t]
\centering
\def\svgwidth{0.9\linewidth}
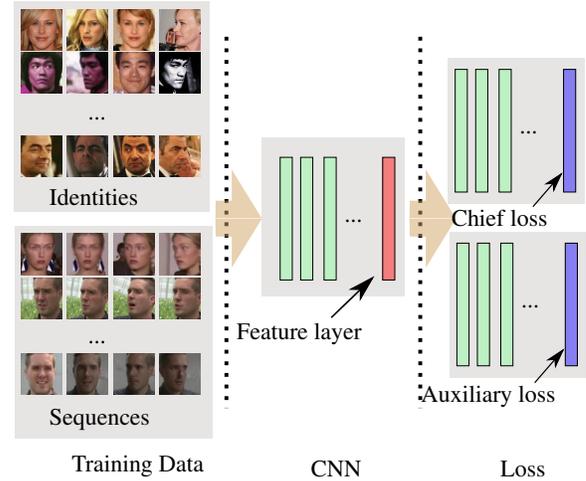
\caption{In our SeqFace framework, the CNN model is trained on an identity dataset and a sequence dataset, and is supervised jointly by a chief classification loss and another auxiliary loss. Different sequences can belong to the same identity in sequence data.}
\label{FigPipeline}
\end{figure}

A sequence dataset includes $M$ faces of $N$ sequences, and each face is labeled as the image of the $i$-th ($0 \leq i < N$) sequence. Because faces in a sequence must belong to one identity (note that different sequences might belong to one identity), sequence data have potential to help to reduce the intra-identity variations. A large-scale high-quality sequence dataset can be efficiently and automatically constructed by using state-of-the-art face detection and tracking methods. Although face sequences are broadly used in video FR applications, previous works have rarely utilized these unlabelled sequence datasets as the training data to learn face features in image FR.

In this paper, we propose a framework, namely SeqFace, to learn discriminative face features on both identity data and sequence data (see Figure~\ref{FigPipeline}). The SeqFace framework is not objective to replace other models or loss functions, but to make full use of sequence data in the training procedure. In the SeqFace, a CNN model is jointly supervised by two loss functions. The first one is a chief classification loss, such as the softmax loss, which aims to maximize the inter-identity variations and minimize the intra-identity variations simultaneously. The second one is an auxiliary loss, such as the center loss~\cite{centerloss}, which mainly encourages the intra-identity compactness. Because the traditional classification loss functions cannot deal with sequence data, label smoothing regularization (LSR) is employed to improve the chief loss. Moreover, we also propose a DSA loss as the auxiliary loss, which is superior to the center loss because it contributes to the inter-class dispersion. With the help of additional sequence data, CNN models can be trained with high feature discrimination in the SeqFace.

To summarize, our major contributions are as follows:

\begin{enumerate}
\item We present a framework (SeqFace) to learn discriminative face features. Besides the traditional identity data, unlabeled sequences are used as the training data in the SeqFace to enhance discrimination power of face features for the first time. 
\item To make full use of sequence data, we employed the LSR to help the chief classification loss deal with sequence data. A new DSA loss function, which contributes to the intra-class compactness and the inter-class dispersion of the features, is also proposed as the auxiliary loss to train CNNs. Experiments demonstrate that the LSR and the DSA loss both boost the FR performance greatly.
\item We conduct experiments on two popular and challenging FR benchmark datasets (Labeled Faces in the Wild (LFW) and YouTube Faces (YTF)~\cite{ytf}) with one single ResNet-64, and the model achieves a 99.83 \% verification accuracy on LFW, and a 98.12\% verification accuracy on YTF.
\end{enumerate}

\section{Related Works}
\label{sec:Related Works}
In this section, we briefly review works of deep face recognition, and face sequences related works in FR are also introduced.

\subsubsection{Deep Face Recognition}
Deep face recognition is one of the most active field, and has achieved a series of breakthroughs in recent years thanks to the great success of CNNs~\cite{resnet,cnn1,cnn2,cnn3}. Many methods~\cite{deepface,frbyclass1,frbyclass2,facenet,deepid2,deepid3,deepid2plus} have  proven that CNNs outperform humans in FR on some benchmark data sets. FR is treated as a multi-class classification problem and CNN models are supervised by the softmax loss in many methods~\cite{deepface,frbyclass1,frbyclass2}. Some metric learning loss functions, such as contrastive loss~\cite{casia,contrastiveloss}, triplet loss~\cite{tripletloss,facenet} and center loss~\cite{centerloss}, are also applied to boost FR performance greatly. Other loss functions~\cite{marginalloss,rangeloss} based on metric loss also demonstrate effective performance on FR. Recently, some angular margin based methods~\cite{lsoftmax,sphereface,arcface,cosface} are proposed and achieve outperforming performance. 

\subsubsection{Sequences in Face Recognition}
In many applications, sequences or image sets are the most natural form of input to the FR system. Video face recognition methods~\cite{facesequence1,facesequence2,facesequence3,videfr1,trunkcnn,coxface,haarnet,unlabeledvideo} based on face sequences, or face sets, also are expected to achieve better performance than ones based on individual images. Most of these studies attempt to utilize redundant information of face sequences/sets to improve recognition performance, but not to learn discriminative features from sequence data. Recently, some approaches~\cite{trunkcnn,haarnet,unlabeledvideo} aim to learn deep video features for video face recognition. In~\cite{unlabeledvideo}, large-scale unlabeled face sequences are employed as the training data, but these sequence data are only utilized to learn transformations between image and video domains. Learning discriminative face features still mainly depends on traditional large-scale identity datasets in this deep CNN approaches of video FR.

\section{The Proposed Approach}
\label{sec:proposed}

\subsection{SeqFace Framework}
The proposed SeqFace is a framework for learning discriminative face features on identity datasets and sequence datasets simultaneously. \iffalse In \emph{identity data}, faces of one identity are labeled as the same $ID_{identity}$. In \emph{sequence data}, faces in one sequence are labeled as the same $ID_{sequence}$. Two faces with the different $ID_{identity}$ must belong to different identities, but two faces with the different $ID_{sequence}$ might (or might not) belong to one identity. \fi Identity overlap between these two datasets is not allowed in the SeqFace. A CNN model (ResNet-like models in our implementation) is jointly supervised by one chief classification loss and one auxiliary loss in the SeqFace. The chief loss enlarges the inter-identity feature differences and reduces the intra-identity feature variations simultaneously, and the major target of the auxiliary loss is to reduce intra-identity(intra-sequence) variations. The final loss can be formulated as
\begin{equation}
\mathcal{L} = \mathcal{L}_{Chief} + \eta \cdot \mathcal{L}_{Auxiliary},
\label{equTotalLoss}
\end{equation}
where $\eta$ is a parameter used to balance two loss functions.

Similar with many methods, we also treat the FR problem as a classification task to train CNNs, and CNNs are mainly supervised by a chief classification loss, such as the Softmax loss, the A-Softmax loss in SphereFace~\cite{sphereface}, and so on. All faces of one identity in \emph{identity data} is labeled as belonging to one class in the classification loss. However, an input face in \emph{sequence data} cannot belong to any class (identity) in the classification loss, because there is no identity annotation in sequence data. That is to say, though a regular classification loss can be applied to train the CNN model in the SeqFace, it only can deal with \emph{identity data} and has to ignore \emph{sequence data}. 

We know that faces in one sequence certainly belong to one identity. Therefore, if a loss encourages the intra-sequence feature compactness, and does not penalize the inter-sequence feature compactness, it could supervise CNNs to learn discriminative face features on \emph{sequence data}, and it could naturally deal with \emph{identity data} too. Because this loss mainly affects the intra-sequence and intra-identity compactness, it has to be an auxiliary loss. The center loss~\cite{centerloss} function is formulated as
\begin{equation}
\mathcal{L}_C = \frac{1}{2}\sum\limits_{k=1}^{K}\|\vec{x}_k-\vec{c}_{y_k}\|_2^2,
\end{equation}
where $K$ is the number of training samples, $\vec{x}_k$ denotes the feature of the $k$-th training sample, $y_k$ is the class(identity) label of the sample, and the $\vec{c}_{y_k}$ denotes the $y_k$-th class(identity) center of deep features for classification problems. The center loss can deals with \emph{identity data} and \emph{sequence data} in the same way, and the $\vec{c}_{y_k}$ is the feature center of the $y_k$-th identity for \emph{identity data} or the feature center of the $y_k$-th sequence for \emph{sequence data}. Since the formulation only characterizes intra-identity and intra-sequence feature compactness effectively, it doesn't penalize closed feature centers of different sequences. Therefore, the center loss is a good option for the auxiliary loss in the SeqFace.

To summarize, one of the benefits of our SeqFace framework is to reduce the intra-identity variations while enlarging the inter-identity variation with CNNs supervised by a chief classification loss and another auxiliary loss. In fact, we can employ the regular softmax loss and the center loss as the chief loss and the auxiliary loss in the SeqFace. However, the softmax loss has to ignore sequence data, and the center loss only concerns the intra-identity and intra-sequence compactness. \emph{Sequence data} only contribute to intra-identity compactness through the supervision of the center loss. In order to make full use of sequence information, the LSR and the DSA loss are presented in the paper.

\subsection{Label Smoothing Regularization}
\label{sec::ssoftmax}

The softmax loss is applied to supervise CNNs classification, and its simplicity and probabilistic interpretation make the softmax loss widely adopted in FR issues. The softmax loss can be used as the chief loss in the SeqFace, but it has to ignore \emph{sequence data} when training CNNs because of the lack of identity annotation. The softmax loss can be considered as the combination of a softmax function and a cross-entropy loss, and the cross-entropy loss is formulated as
\begin{equation}
\label{equLoss}
\mathcal{L}_S=-\sum\limits_{i=1}^Clog(p(i))q(i),
\end{equation}
where $C$ is the class number, $p(i) \in [0,1]$ is the predicted probability (the output of the softmax function) of the input belonging to class $i$, and $q(i)$ is the ground truth distribution defined as
\begin{equation}
\label{equQK}
q(i)=
\begin{cases}
0& i \neq y\\
1& i = y
\end{cases},
\end{equation}
where $y$ is the ground truth class label of the input. Label smoothing regularization (LSR) is introduced to deal with non-ground truth inputs~\cite{lsr}, and label smoothing regularization for outlier (LSRO) is then used to corporate unlabeled inputs~\cite{lsro} in CNNs. In the LSR, the value of $q(i)$ can be a float value between 0 and 1 for the input which cannot be clearly labeled as any class.

In our framework, because all identities in \emph{sequence data} do not exist in \emph{identity data}, we define $q(i) = 1/C$ (so $\sum\nolimits_{i=1}^Cq(i)=1$) as~\cite{lsro} for all input faces in \emph{sequence data}. Therefore, the cross-entropy loss is rewritten as
\begin{equation}
L=-(1-Z)log(p(y))-\frac{Z}{C}\sum\limits_{i=1}^Clog(p(i)),
\end{equation}
where $Z=0$ for the input face of \emph{identity data}, and $Z=1$ for the input face of \emph{sequence data}.

The LSR can also be integrated into other softmax-like classification loss functions. In practice, a feature normalised SphereFace (L2-SphereFace for short, the same with FNorm-SphereFace in~\cite{arcface}) is applied as the chief classification loss. An additional $L_2$-constraint is added to the regular SphereFace~\cite{sphereface}, it means the input feature $\vec{x}_k$ must be firstly normalized and scaled by a scalar parameter $\delta$ ($\delta \cdot \vec{x}_k / \|\vec{x}_k\|_2$). Therefore, the decision boundaries of the L2-SphereFace under binary classification is $\delta(\cos m\theta_1-\cos \theta_2)=0$ for class 1, and is $\delta(\cos \theta_1-\cos m\theta_2)=0$ for class 2. In our implementation, the parameter $\delta$ and the margin $m$ are set to 32.0 and 4 respectively.

\subsection{DSA loss}
\label{secDSA}

The center loss only reduces intra-class variations as the auxiliary loss, and the inter-class separability of features completely depends on the classification loss. In this section, we propose a new auxiliary loss, namely discriminative sequence agent loss (DSA Loss), which concerns the intra-class compactness and the inter-class dispersion simultaneously.

First, considering the traditional classification problem with an identity dataset, we define 
\begin{equation}
d_{k,n}=\frac{(\vec{x}_k - \vec{c}_n)^2}{4}
\label{equDistance}
\end{equation}
 as the distance between the feature $\vec{x}_k$ of the $k$-th training sample and the feature center $\vec{c}_n$ of the $n$-th class(identity), $d_{k,n}$ is actually equivalent to the Euclidean distance. Note that if $\vec{x}_k$ and $\vec{c}_n$ are normalized, $d_{k,n}$ can be re-formulated as
\begin{equation}
d_{k,n} = \frac{(1-cos \theta_{k,n})}{2},
\end{equation} 
where $\theta_{k,n}$ denotes the angle between $\vec{x}_k$ and $\vec{c}_n$, and $d_{k,n}$ can be regarded as the angular distance. Since our target is to reduce the distance between $\vec{x}_k$ and $\vec{c}_{y_k}$ and enlarge other distances between $\vec{x}_k$ and $\vec{c}_{n}$ for all $n \neq y_k$, where $y_k$ is the label of the $k$-th training sample, a discriminative loss can be formulated as 
\begin{equation}
\mathcal{L}_{k,n}=
\begin{cases}
d_{k,n} & n = y_k\\
max(\alpha \cdot d_{k,y_k} - d_{k,n} + \beta, 0) & n \neq y_k
\end{cases},
\end{equation}
where $\alpha \in [1,+\infty)$ and $\beta \in [0,+\infty)$ are two parameters to adjust the discriminative power of the learned features. Therefore, the final loss function is
\begin{equation}
\mathcal{L}_{D}=\frac{1}{K}\sum \limits_{k=1}^{K} \left[ \lambda \cdot \mathcal{L}_{k,y_k} + 
(1 - \lambda )\frac{1}{(N-1) \cdot p} \sum \limits_{n=1, n \neq y_k}^{N}b(1,p) \cdot \mathcal{L}_{k,n} \right],
\label{equDSALoss}
\end{equation}
where the parameter $\lambda$ is applied to balance the intra-class compactness and the inter-class dispersion, $N$ is the number of identity(class) of the identity dataset. We introduce another parameter $p$ as the probability that the $n$-th center is employed in computing the final loss, because $N$ might be a huge number and it will be time-consuming if all $\mathcal{L}_{k,n}$ are computed in each iteration. $b(1,p)$ means the Bernoulli distribution with the probability $p$. 

The gradients of $\mathcal{L}_{D}$ with respect to $\vec{x}_k$ and the update equation of $\vec{c}_n$, similar with that in the center loss, are computed as:
\begin{equation}
\begin{split}
\frac{\partial \mathcal{L}_{D}}{\partial \vec{x}_k}= & \frac{1}{K} \biggr[ \lambda \cdot \frac{\vec{x}_k - \vec{c}_{y_k}}{2} +  (1 - \lambda )\frac{1}{(N-1) \cdot p} \cdot \\ & \sum \limits_{n=1, n \neq y_k}^{N}b(1,p) \cdot  \delta (L_{k,n}>0) \cdot 
 (\alpha \cdot \frac{\vec{x}_k - \vec{c}_{y_k}}{2} - \frac{\vec{x}_k - \vec{c}_{n}}{2}) \biggr]
\end{split}
\end{equation}
and 
\begin{equation}
\Delta \vec{c}_n = - \frac{\sum \nolimits_{n=1}^N \delta (y_k = n) \cdot \frac{\vec{x}_k - \vec{c}_n}{2}}{ 1 + \sum \nolimits_{n=1}^N \delta (y_k = n) },
\end{equation}
where $\delta (condition) = 1$ if the $condition$ is satisfied, and $\delta (condition) = 0$ if not.

Different from the center loss, our DSA loss also enforces constraints on inter-class variations. According to Equation~\ref{equDSALoss}, the feature $\vec{x}_k$ is pulled towards the feature center $\vec{c}_{y_k}$ of its identity, and is pushed away from feature centers of other identities randomly selected in each training iteration. 
\begin{figure}[htb]
\centering
\def\svgwidth{0.60\linewidth}
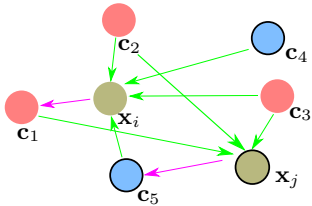
\caption{Illustration of \emph{forces} on sample features of identity data and sequence data. The $i$-th sample is from the identity dataset, and the $j$-th one is from the sequence dataset. $\vec{c}_1$, $\vec{c}_2$ and $\vec{c}_3$ are feature centers of corresponding identities, and $\vec{c}_4$ and $\vec{c}_5$ are feature centers of corresponding sequences. $y_i=1$ and $y_j=5$.}
\label{FigDSALoss}
\end{figure}

Taking into account \emph{sequence data}, there is a slight modification in Equation~\ref{equDSALoss} to compute the final DSA loss. We assume that there are $C$ ($C$ is also the number of class in the chief classification loss) identities in the identity dataset and $N-C$ sequences in the sequence dataset. There is no overlap between these two datasets as mentioned above. In Equation~\ref{equDSALoss}, $n$ is selected from all $C$ identities and $N-C$ sequences (means $\sum\nolimits_{n=1,n\neq y_k}^{N}$) for samples in the identity dataset, and $n$ is only selected from $C$ identities (means $\sum\nolimits_{n=1,n\neq y_k}^{C}$) for samples in the sequence dataset. That is to say, if the $k$-th sample is in the identity dataset, $\vec{x}_k$ should be pushed away from feature centers of other identities and all sequences, or $\vec{x}_k$ is only pushed away from feature centers of identities. Figure~\ref{FigDSALoss} illustrates two examples.

There are four parameters ($\lambda$, $\alpha$, $\beta$, and $p$) in the DSA loss function. The parameter $\lambda$ can be set to 0.5 since we concern both the intra-class compactness and the inter-class dispersion. The parameters $\alpha$ and $\beta$ are used to adjust the discriminative power of features. Using larger values is preferred, but it will increase the difficulty of convergence in training. According to our experiments, $\alpha=2.0$ and $\beta=1.0$ can be applied in most applications. The parameter $p$ is applied to select part of identities/sequences while computing $\mathcal{L}_{k,n}$, in order to reduce the computing cost. The value of the parameter $p$ can be set flexibly based on computing resources in real applications. 

\subsubsection{MNIST Example} 
\begin{figure*}[htb]
\centering
\subfigure[]
{\includegraphics[width=0.23\linewidth]{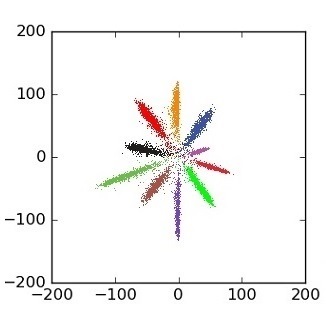}}
\subfigure[]
{\includegraphics[width=0.23\linewidth]{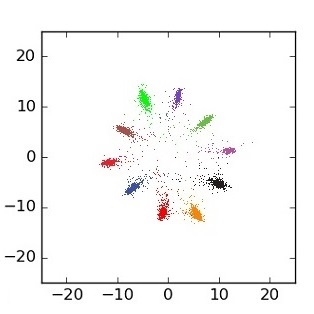}}
\subfigure[]
{\includegraphics[width=0.23\linewidth]{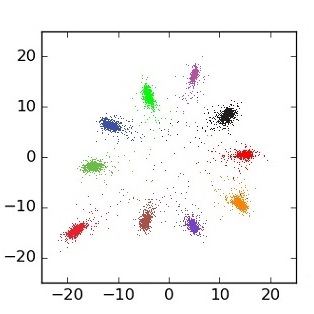}}
\subfigure[]
{\includegraphics[width=0.23\linewidth]{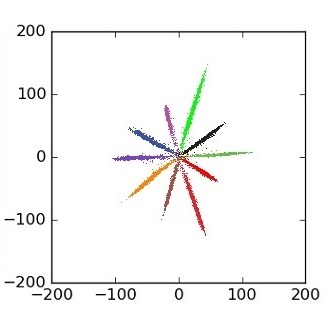}}
\caption{Visualization of 2-D feature distribution for the MNIST \emph{test set}. The features of samples from different classes are denoted by the points with different colors. Four CNNs are supervised by the loss functions of (a)Softmax loss. (b)Softmax loss + Center loss. (c)Softmax loss + DSA loss with Euclidean distance. (d)Softmax loss + DSA loss with angular distance.}
\label{figMNIST}
\end{figure*}

We perform a toy example on the MNIST dataset~\cite{mnist} with our DSA loss. LeNet++~\cite{centerloss}, a deeper and wider version of LeNet, is employed. The last hidden layer output of the model is restricted to 2-dimensions for easy visualization (see Figure~\ref{figMNIST}). For comparison, we train 4 models supervised by a softmax loss, a softmax loss and a center loss, a softmax loss and a DSA loss, a softmax loss and a DSA loss (with normalized $\vec{x}_k$ and $\vec{c}_n$), respectively. We set $\lambda=0.5$, $\alpha=2.0$, $\beta=1.0$ and $p = 1.0$ in the DSA loss. The loss weight values of the center/DSA loss are set to 0.04. All models are trained with the batch size of 32. The learning rate begins with 0.01, and is divided by 10 at 14K iterations. The training process is finished at 20K iterations. As shown in Figure~\ref{figMNIST}, the features learned with the DSA loss are more discriminative. The feature dispersion in Figure~\ref{figMNIST}(b) and Figure~\ref{figMNIST}(c) demonstrates that the DSA loss can enlarge inter-class distances, and the feature centers of different classes are pushed away from each other.

\iffalse Table~\ref{tab_mnist} lists the classification accuracies of 4 models on the test set. From the results, we can have following observations: 1) The center loss and the DSA loss both improve the classification performance; 2) Our DSA loss outperforms the center loss as a auxiliary loss.\fi
\iffalse
\begin{table}
\caption{Accuracy on MNIST test set}
\centering
\begin{tabular}{cc}
\hline
Loss & Accuracy\\
\hline
Softmax & 98.86\%\\
Softmax + Center & 99.06\%\\
Softmax + Euclidean distance DSA & 99.17\%\\
Softmax + Angular distance DSA & 99.23\%\\
\hline
\end{tabular}
\label{tab_mnist}
\end{table}
\fi

\section{Experiment}
\label{sec:Experiment}
\subsection{Implement Details}

In our experiments, all the face images and their landmarks are detected by MTCNN~\cite{mtcnn}. The faces are aligned by similar transformation as~\cite{lightcnn}, and are cropped to $144 \times 144$ RGB images (randomly cropped to $128 \times 128$ in training). Each pixel in RGB images is normalized by subtracting 127.5 then divided by 128. 

\subsubsection{Training and Testing} Caffe~\cite{caffe} is used to implement CNN models. Different CNN models are employed in the experiments, which will be further introduced. All weights of the auxiliary losses ($\eta$ in Equation~\ref{equTotalLoss}) are set to 0.04 in the experiments. Euclidean distances (do not normalize $\vec{x}_k$ and $\vec{c}_n$ in Equation~\ref{equDistance}) are applied in the DSA loss functions used in these section. At the testing stage, \textbf{only features of the original image} are directly extracted from the last full connected layer of CNNs, and the cosine similarity is used to measure the feature distance in the experiments. More details are presented in the corresponding sections. 

\subsection{Exploration Experiment}
\label{secExp}
In this section, the employed CNN is a ResNet-20 network which is similar to~\cite{sphereface}, and it is trained on the publicly available CASIA-WebFace dataset~\cite{casia} containing about 0.5M faces from 10,575 identities. All models are trained with the batch size of 32 on one Titanx GPU. The learning rate begins with 0.01, and is divided by 10 at 200K iterations. The training process is finished at 300K iterations.  

To evaluate the effectiveness of sequence data, 10,575 identities in the CASIA-WebFace dataset are randomly divided into two parts: the dataset A (5,000 identities) and the dataset B(5,575 identities). Faces in the dataset B is then randomly split into 32,996 sequences. The dataset A and B are treated as the identity dataset and the sequence dataset respectively.
\begin{figure}[htb]
\centering
{\includegraphics[width=0.85\linewidth]
{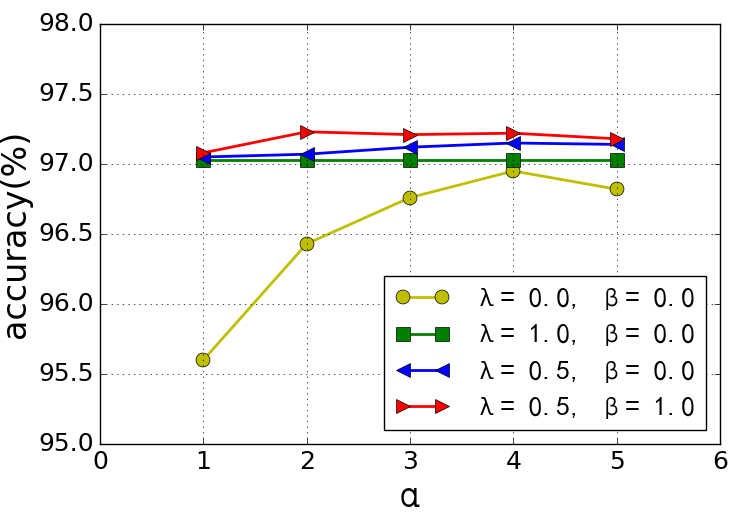}}
\caption{Face verification accuracies on LFW achieved by the DSA losses with different parameter values. }
\label{figDSAParam}
\end{figure}
\subsubsection{Effect of the DSA loss parameters} We firstly study the effect of parameter values in the DSA loss. We train several CNN models only on the dataset A(5,000 identities) under the supervision of the DSA loss with different parameter values, and then evaluate these models on the LFW dataset. From the results shown in Figure~\ref{figDSAParam}, we can conclude that higher performances can be achieved if the DSA loss simultaneously concern the intra-class compactness and the inter-class dispersion of the learned features ($\lambda=0.5$). The results also demonstrate that larger $\alpha$ and $\beta$ might lead to more discriminative features. According to our experiments, setting $\alpha=2.0$ and $\beta=1.0$ is preferred to balance the FR performance and the difficulty of convergence.

\iffalse The effectiveness of the DSA loss is demonstrated from the mnist example in Section~\ref{secDSA}. To further explore its effectiveness on FR issues, we train 3 CNN models under the supervision of a softmax loss, a softmax loss + a center loss, a softmax loss + a DSA loss on the dataset A, and then evaluate these models on the LFW dataset. We set $\lambda=0.5$, $\alpha = 2.0$, $\beta = 1.0$ and $p = 0.01$ in the DSA loss. The DSA loss based on Euclidean distance is applied for comparing with the center loss. From the results shown in Table~\ref{tab_dsa}, we can see that jointly supervision can notably enhance the discriminative power of face features, and the DSA loss achieves better performance than the center loss.
\begin{table}
\caption{Face verification accuracies on LFW}
\centering
\begin{tabular}{cc}
\hline
Loss & Accuracy\\
\hline
Softmax & 95.12\%\\
Softmax + Center & 97.03\%\\
Softmax + DSA & 97.25\%\\
\hline
\end{tabular}
\label{tab_dsa}
\end{table}
\fi
\subsubsection{Effect of SeqFace} We also train 10 models (see Table~\ref{tab_seqface}) to demonstrate the effectiveness of our SeqFace, the LSR and the DSA loss.

First, we use a regular softmax loss (Model \Rmnum{1}), a L2-SphereFace loss (Model \Rmnum{2}) to train 2 CNN models respectively. \textbf{Only the dataset A} is used as the training dataset. Verification accuracies demonstrate that the L2-SphereFace greatly boosts the performance.

Then, the center loss and the DSA loss are applied as the auxiliary loss to jointly supervise 2 CNN models (Model \Rmnum{3} and Model \Rmnum{4}) with a softmax loss respectively. The reported results also demonstrate that: 1) Auxiliary loss functions have positive effect on the FR performance even without sequence training data; 2) Our DSA loss outperforms the center loss on FR issue.

Moreover, \emph{sequence data} (\textbf{the dataset B}) are added to train 5 CNN models supervised by a LSR-based softmax loss and a center loss (Model \Rmnum{5}), a LSR-based softmax loss and a DSA loss (Model \Rmnum{6}), a LSR-based L2-SphereFace loss (Model \Rmnum{7}), a LSR-based L2-SphereFace loss and a center loss (Model \Rmnum{8}), a LSR-based L2-SphereFace and a DSA loss (Model \Rmnum{9}), respectively. According to results, we have following observation: 1) Sequence data can obviously improve the FR performance on the SeqFace; 2) Even one chief classification loss with the LSR can also utilize sequence data to improve the FR performance; 3) The LSR and the DSA loss greatly enhance the discriminative power of learned features.

Last, we also train a model (Model \Rmnum{10}) on total CASIA-WebFace with a L2-SphereFace loss, and it reaches 99.03\% accuracy on LFW. Comparing accuracies between the Model \Rmnum{9} and \Rmnum{10}, we can conclude that complete identity annotation is naturally preferred in training datasets, but the little gap shows that competitive performance also can be achieved by making full use of sequence information.
\begin{table*}[t]
\caption{Face verification accuracy on LFW dataset}
\centering
\begin{tabular}{c|ccc}
\hline
Model & Loss & Training Dataset & Accuracy\\
\hline
\Rmnum{1} & Softmax loss & Dataset A & 95.12\%\\
\Rmnum{2} & L2-SphereFace & Dataset A & 97.35\%\\
\Rmnum{3} & Softmax + Center loss & Dataset A & 97.03\%\\
\Rmnum{4} & Softmax + DSA loss & Dataset A & 97.25\%\\
\Rmnum{5} & LSR-Softmax + Center loss & Dataset A + Dataset B & 97.62\%\\
\Rmnum{6} & LSR-Softmax + DSA loss & Dataset A + Dataset B & 98.13\%\\
\Rmnum{7} & LSR-L2-SphereFace & Dataset A + Dataset B & 98.65\%\\
\Rmnum{8} & LSR-L2-SphereFace + Center loss & Dataset A + Dataset B & 98.72\%\\
\Rmnum{9} & LSR-L2-SphereFace + DSA loss & Dataset A + Dataset B & 98.85\%\\
\Rmnum{10} & L2-SphereFace & CASIA-WebFace & 99.03\%\\
\hline
\end{tabular}
\label{tab_seqface}
\end{table*}

\subsection{Evaluation on LFW and YTF}
\label{secBenchmark}

In this section, we evaluate the proposed SeqFace on LFW and YTF in unconstrained environments. LFW~\cite{lfw} and YTF~\cite{ytf} are challenging testing benchmarks released for face verification. LFW dataset contains 13,233 faces of 5749 different identities, with large variations in pose, expression and illuminations. YTF dataset includes 3,425 videos of 1,595 identities. We follow the unrestricted with labeled outside data protocol. To evaluate performance on YTF, the simple average feature of all faces in a video is applied to compute the final score.

\begin{figure*}[htb]
\centering
\def\svgwidth{0.99\linewidth}
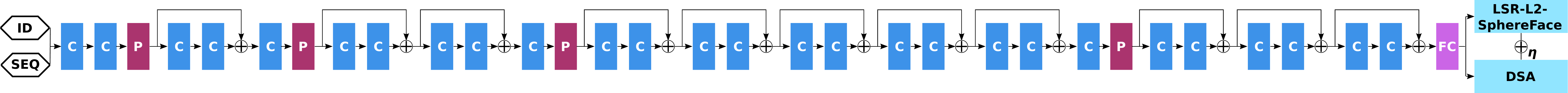
\caption{The ResNet-27 architecture for the experiments on LFW and YTF. The CNN is jointly supervised by the LSR-L2-SphereFace and the DSA loss. \textbf{ID} denotes identity input data, \textbf{SEQ} denotes sequence input data, \textbf{C} denotes the convolution layer, \textbf{P} denotes the max-pooling layer, and \textbf{FC} denotes the fully connected layer.}
\label{figCNN}
\end{figure*}

A ResNet-27 model\footnote{https://github.com/ydwen/caffe-face}(the architecture is shown in Figure~\ref{figCNN}) and a ResNet-64~\cite{sphereface} are employed for evaluation. To accelerate the training process, we first train a baseline model under the supervision of the regular L2-SphereFace on the identity dataset only, and then fine-tune the baseline model by using the SeqFace. Our models are trained with batch size of 128 on 4 Titanx GPU. The learning rate begins with 0.01, and is divided by 10 at 300K and 600K iterations. The training is finished at 800K iterations. The model is jointly supervised by a LSR-L2-SphereFace loss and a DSA loss, and is learned on the MS-Celeb-1M and our Celeb-Seq datasets described below. In the DSA loss, $\lambda=0.5$, $\alpha = 2.0$, $\beta = 1.0$. The parameter $p$ is set to 0.001 because of the large number of sequences in the Celeb-Seq dataset.

\subsubsection{Training Datasets} 
\begin{figure*}[htb]
\centering  
\def\svgwidth{0.95\linewidth}
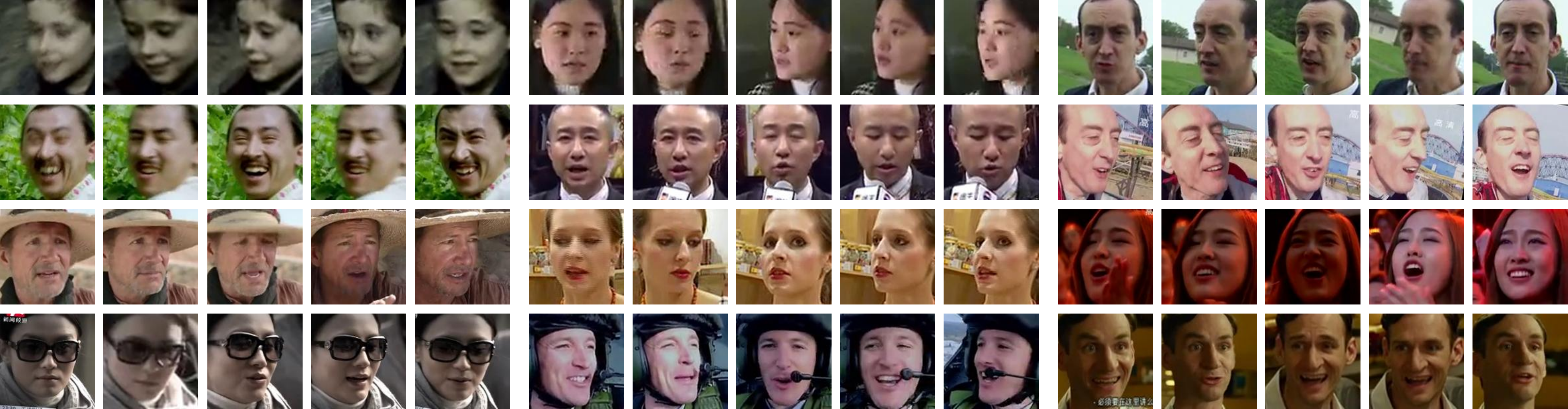
\caption{Some face sequences in the Celeb-Seq dataset. All faces are aligned and cropped to $144 \times 144$. Some sequences belong to the same identity (two sequences at top-right corner). Note that the numbers of faces in the sequences are different from each other.}
\label{figCeleb}
\end{figure*}

A refined MS-Celeb-1M (4M images and 79K identities) provided by~\cite{lightcnn} is used as the identity dataset. Since there is no public sequence datasets for training deep CNNs, we construct a sequence dataset Celeb-Seq, which includes about 2.5M face images of 550K face sequences. We firstly extract about 800K face sequences by using MTCNN~\cite{mtcnn} and Kalman-Consensus Filter (KCF)~\cite{kcf} to detect and track video faces from 32 online TV Channels, then compute image features with the model provided by SphereFace~\cite{sphereface}. Lastly faces of overlap identities with MS-Celeb-1M, and nearly noisy and duplicate faces in one sequence are discarded from the dataset automatically and manually. We also remove face images belong to identities that appear in the LFW and YTF test sets. Some face sequences in the Celeb-Seq dataset are shown in Figure~\ref{figCeleb}.

Removing overlap identities with MS-Celeb-MS costs us most of the time in the constructing process, because many celebrities in MS-Celeb-MS can be found from the original 800K sequences. Fortunately, constructing a satisfied sequence dataset will not be a time-consuming task in many real scenarios. For example, it is almost certain that people in an Asian street surveillance video will not appear in another European street surveillance video, and will not be found in the MS-Celeb-1M dataset too.

\subsubsection{Evaluation Results}
Table~\ref{tab_final} reports the verification performance of several methods. To demonstrate effectiveness of the SeqFace, the performance of our baseline ResNet-27 is also reported in the table. Note that the ArcFace employs the improved ResNets~\cite{newresnet}. The SeqFace achieves the best accuracies on these two benchmark testing sets. It is reported in the ArcFace that a regular 50-layer ResNet achieves a 99.71\% accuracy on LFW. Moreover, our ResNet-27 and ResNet-64 models achieve \textbf{99.50\%} and \textbf{99.67\%} at VR@FAR=0 on LFW.

The SeqFace is only a framework to make use of sequence data. We believe that new loss functions (such as ArcFace, CosFace, etc.) and a deeper ResNet with improved residual units can be employed in the SeqFace to further improve the performance.  

\begin{table*}[t]
\caption{Verification accuracies(\%) of different methods on LFW and YTF. Note that ResNet models in the ArcFace used the improved residual units, and the training MS-Celeb-1M dataset used in the ArcFace contains 3.8M images and 85K identities.}
\centering
\setlength{\tabcolsep}{1mm}{
\begin{tabular}{c | c c  c  c}
\hline
Method & Models & Data & LFW & YTF\\
\hline
DeepFace~\cite{deepface} & 3 & 4M images & 97.35 & 91.4 \\
DeepID2+~\cite{deepid2plus} & 1 & 300K images & 98.70 & - \\
DeepID2+~\cite{deepid2plus} & 25 & 300K images & 99.47 & 93.2 \\
FaceNet~\cite{facenet} & 1 & 200M images & 99.65 & 95.1 \\
Baidu~\cite{baidu} & 9 & 1.2M images & 99.77 & - \\
\hline
Center Face~\cite{centerloss} & 1 & 0.7M images & 99.28 & 94.9 \\
SphereFace~\cite{sphereface} & 1 ResNet-64 & CASIA-Webface & 99.42 & 95.0 \\
CosFace~\cite{cosface} & 1 ResNet-64 & 5M images & 99.73 & 97.6 \\
ArcFace~\cite{arcface} & 1 ResNet-50 & MS-Celeb-1M & 99.78 & - \\
ArcFace~\cite{arcface} & 1 ResNet-100 & MS-Celeb-1M & \textbf{99.83} & - \\
\hline
L2-SphereFace & 1 ResNet-27 & MS-Celeb-1M & 99.55 & 95.7 \\
SeqFace & 1 ResNet-27 & MS-Celeb-1M + Celeb-Seq & 99.80 & 98.0 \\
SeqFace & 1 ResNet-64 & MS-Celeb-1M + Celeb-Seq & \textbf{99.83} & \textbf{98.12} \\
\hline
\end{tabular}}
\label{tab_final}
\end{table*}
\section{Conclusion}
\label{sec:conclusion}
A large-scale high-quality dataset for training CNNs in FR is very expensive to construct. Face features learned on publicly available datasets for researchers might not achieve satisfied performance in some circumstances, for example evaluating Asia people in surveillance videos. Though large amount of face images in the real situation can be collected, assigning labels to these images is still time-consuming. Fortunately, a dataset containing large amount of face sequences can be efficiently constructed by using face detection and tracking methods.

In this paper, we proposed a framework named SeqFace, which can utilize sequence data to learn highly discriminative face features. A chief classification loss and another auxiliary loss are combined together to learn features on a traditional identity dataset and another sequence dataset. The LSR is employed to help the chief loss to deal with sequence input. The DSA loss was also proposed to supervise CNNs as an auxiliary loss. We achieved good results on several popular face benchmarks only with a simple ResNet model. We also believe that more competitive performance can be obtained, if recently proposed loss functions~\cite{cosface,arcface} and CNN architectures~\cite{newresnet} are employed.

As far as we know, SeqFace is the first framework to employ face sequences as training data to learn highly discriminative face features. The requirement of no identity overlap between the identity and sequence datasets might have influence on the efficiency of constructing sequence datasets, but it always happens in many situations mentioned above. In fact, we train a CNN model on MS-Celeb-1M dataset and another sequence dataset, whose face sequences are collected from surveillance videos in China. The learned model achieves good performance in surveillance applications. It is obvious that our SeqFace also has great potentiality to be applied in other similar fields, such as Person-reidentification.

%\section*{Acknowledgements}

\bibliographystyle{acmsiggraph}
%\bibliography{seqface}

\end{document}